# Table Tennis Stroke Recognition Using Two-Dimensional Human Pose Estimation

Kaustubh Milind Kulkarni
kmkapril15@gmail.com

Sucheth Shenoy
sucheth17@gmail.com

**Abstract**

*We introduce a novel method for collecting table tennis video data and perform stroke detection and classification. A diverse dataset containing video data of 11 basic strokes obtained from 14 professional table tennis players, summing up to a total of 22111 videos has been collected using the proposed setup. The temporal convolutional neural network model developed using 2D pose estimation performs multiclass classification of these 11 table tennis strokes with a validation accuracy of 99.37%. Moreover, the neural network generalizes well over the data of a player excluded from the training and validation dataset, classifying the fresh strokes with an overall best accuracy of 98.72%. Various model architectures using machine learning and deep learning based approaches have been trained for stroke recognition and their performances have been compared and benchmarked. Inferences such as performance monitoring and stroke comparison of the players using the model have been discussed. Therefore, we are contributing to the development of a computer vision based sports analytics system for the sport of table tennis that focuses on the previously unexploited aspect of the sport i.e., a player's strokes, which is extremely insightful for performance improvement.*

## 1. Introduction

Performance monitoring is a very essential aspect of professional sports training. It helps in constantly tracking the progress of a sportsperson and also helps to determine the areas which have scope for improvement. The traditional methods of sports training involving just players and coaches have saturated over the years and hence, there exists a dire need for the incorporation of technology in this field. Computer vision based sports analytics systems are being developed for this purpose to provide effective means for collecting characteristic data pertaining to the sport, draw meaningful insights from the data, and provide user-friendly representations of the inferences made. These inferences can be used for continuous improvement in the performance level of a sportsperson which might not be possible using traditional training methods.

Table tennis is an intricate sport involving a wide variety of strokes. These strokes when analysed can provide helpful inferences which in turn can lead to an improvement in the performance of a player. We propose a start in this process of stroke analysis by first collecting the appropriate dataset and developing a neural network which recognizes the table tennis stroke executed by a player, classifying it into one of eleven major strokes.

The strokes executed by a player are clearly visible to the opponent and hence, valuable inferences about the strokes can easily be made from the front view of the player. A major challenge faced was to collect a suitable video dataset as there were no public datasets available for the purpose of table tennis stroke recognition or from which it could be effectively performed. Therefore, we collected a diverse dataset using a novel setup, which has minimal interference with the players as well as the sport.

Different strokes are executed by varying the trajectory of the table tennis racket with respect to the ball which leads to the fact that the nature and range of motion of the arm vary for each kind of stroke. Hence, we propose a method wherein we use two-dimensional human pose estimation to recognize a player's stroke.

## 2. Related works

Table tennis is a popular sport with a wide corpus of strokes. This has attracted plenty of researchers to work in the field of stroke detection and classification for the sport. The initial attempts in this direction are mainly comprised of sensor-based approaches. For example, in the work carried out by Peter Blank *et al.* [1], Inertial Sensors were attached to table tennis rackets to collect stroke data. Strokes were detected using an event detection algorithm with a sensitivity of 95.7% and a classifier used for stroke classification provided an accuracy of 96.7% for 8 basic strokes. Liu R. *et al.* [2] have employed Body Sensor Networks (BSN) to collect motion data pertaining to the upper arm, lower arm and the back of players executing strokes. They obtained an accuracy of 97.41% using Principal Component Analysis (PCA) over a Support Vector Machine (SVM) to recognise the strokes. A comparative study of table tennis forehand strokes has been done by SS Tabrizi *et al.* [3] between SVM with Radial Basis Function (RBF) as its kernel, Long Short-Term Memory (LSTM) and two-dimensional Convolutional

Neural Network (2D-CNN) models using the signals generated by a BNO055 sensor during the execution of forehand table tennis strokes.

The application of computer vision based approaches on table tennis data is a recent trend and there is a scarce amount of work done in the area of stroke recognition. IR depth camera has been used by Habiba Hegazy et al. [4] for the detection and efficiency classification of strokes. An average accuracy of 88% to 100% was obtained by them using the fastDTW algorithm. A Twin Spatio-Temporal Convolutional Neural Network (TSTCNN) based approach for action recognition in table tennis was introduced by Pierre-Etienne Martin et al. [5]. It performed stroke detection and recognition on the TTStroke-21 dataset with an accuracy of 91.4%. Roman Voelikov et al. [6] presented a neural network, TTNet, for real-time temporal and spatial event spotting for table tennis. They obtained a 97.0% accuracy in event spotting and an accuracy of 97.5% in ball detection on the test part of their multi-task dataset, OpenTTGames. Another multi-task dataset for tracking and action recognition in table tennis, SPIN, was presented by Steven Schwarcz et al. [7], where pose tracking and spin prediction have been explored.

## 3. Dataset collection

Although datasets exist for table tennis video data, they did not provide an optimum viewing angle to observe a stroke with high detail. Our dataset has been collected in a way that it provides the most optimum view of a player to break down and analyse the stroke executed by the player. Hence, we developed a system for video data collection which automated the labelling process for the entire dataset. Tab. 1 and Tab. 2 show a detailed description of the dataset collected.

|    | Class              | Number of videos (strokes executed) |
|----|--------------------|-------------------------------------|
| 1  | Forehand Topspin   | 2003                                |
| 2  | Backhand Topspin   | 2403                                |
| 3  | Forehand Push      | 2082                                |
| 4  | Backhand Push      | 2066                                |
| 5  | Forehand Block     | 2019                                |
| 6  | Backhand Block     | 2025                                |
| 7  | Forehand Flick     | 2068                                |
| 8  | Backhand Flick     | 2048                                |
| 9  | Forehand Lob       | 1811                                |
| 10 | Backhand Lob       | 1795                                |
| 11 | Forehand Flat      | 1791                                |

Table 1: Stroke-wise distribution of the dataset.

| Total Number of Strokes | Total Number of Frames | Average Number of Frames (per stroke) | Total Number of Data Points Collected |
|---|---|---|---|
| 22,111 | 873,028 | 39.48 | 6,984,224 |

Table 2: Statistics of the dataset.

### 3.1. Setup

The camera is positioned facing the player, right in front of the net and parallel to it. This enables the camera to capture videos of the player during stroke execution from the front view. To facilitate this, a wooden frame was designed as shown in Fig. 1 to be placed on the table as depicted in Fig. 2 such that it has least interference with regular game play. The height of the net above the table tennis table according to the rules of the sport is 152.5 mm. The total height of the frame is 135 mm so that it does not protrude above the height of the net. A Raspberry Pi Camera Module interfaced with a Raspberry Pi 4 are both mounted onto the rear side of the frame to eliminate the chances of ball impact with the hardware. The camera aperture is at a height of 105 mm above the table surface. The width of the frame is 100 mm providing sufficient clearance for the mounting of the Raspberry Pi 4 and the maximum thickness of the frame is 25 mm at the base.

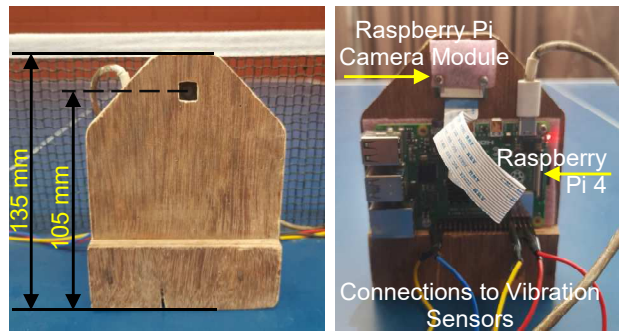

Figure 1: Front view and back view of the camera frame.

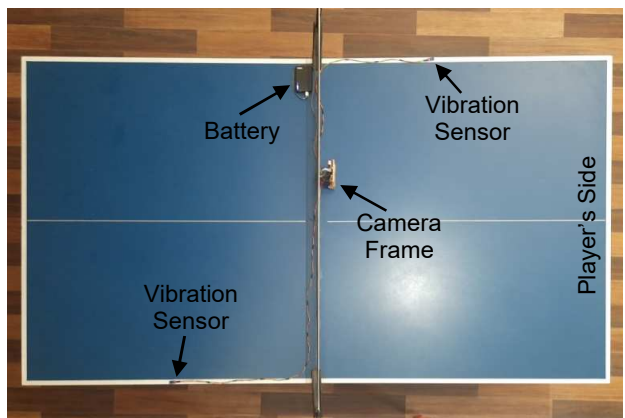

Figure 2: Top view of our setup for dataset collection.

A frame captured by the camera using the setup is depicted in Fig. 3. It captures the entire upper torso of the player executing the stroke along with a small extent of the edge of the table at the bottom of the frame.

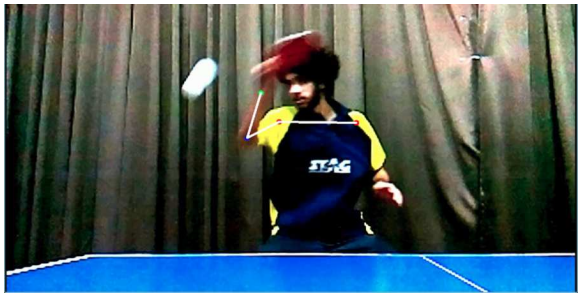

Figure 3: A frame captured using the setup with the outputs of pose estimation mapped for right wrist, right elbow and shoulder points of the player during stroke execution.

Generally, a table tennis player prepares for the consecutive stroke immediately after the execution of the previous stroke. The preparation stage consists of the movement of the player towards the anticipated incoming ball pitch location and the backswing phase of the stroke. This stage contains very less features that can be used for stroke recognition. The next stage of the stroke begins post ball-pitch on the table and it includes the forward stroke phase until ball-contact with the racket and the follow through post ball-contact, which continues approximately till the ball pitches on the other side of the table. These two phases of the stroke are distinct for different strokes and hence provide a high density of features that are necessary for stroke recognition. The final phase of the stroke is the return to neutral position which does not contain any useful features for our purpose. Therefore, video data collected in the time duration between the pitch of the ball on the camera side of the table and the pitch of the ball on the opposite side provides sufficient features to carry out stroke recognition. To automate this aspect of video data capturing, we used two SW-420 digital vibration sensor modules, one mounted on each side of the table.

### 3.2. Procedure and execution

The sensitivities of the vibration sensor modules mounted on each side of the table were calibrated to detect ball pitch on the respective side of the table without producing any false positives or negatives. The camera was set to record videos with an image resolution of width=1280 and height=720 at 60 frames per second (fps).

Data collection and labelling was automated based on the stroke executed, dominant hand of the player in-frame, and number of strokes executed. The vibrations sensors were triggered by the pitching of the ball, providing the opportunity to collect data at a rapid pace as the professional players of the dataset carried on at their natural rally pace, without having to change any aspect of their game to aid the data collection process. The vibration sensor on the player's side of the table, when triggered, initiated recording of the video, and the vibration sensor on the other side, when triggered, stopped the recording of the video. Thus, each video contained a player executing one single stroke. The 11 strokes considered in our dataset were topspin, block, push, flick and lob in both their forehand and backhand variants, and forehand flat.

Data collection for each stroke was made as efficient as possible by recording only the stroke action in the forward stroke and follow through phases. The setup also eliminated the need for manual labelling or trimming of the video data collected. Blocks, pushes, lobs, and forehand flat were collected by allowing the two players to have a continuous free-flowing rally. This accelerated our process of data collection for these strokes. For topspins and flicks, we adopted the multi-ball approach of collecting data. This approach involves an individual feeding table tennis balls to a player on the opposite end of the table. Strokes are executed by the player without any return from the opposite side. This helped reduce the errors made by the player executing the stroke and thus, accelerated our process of data collection for technical strokes.

### 3.3. Dataset preparation and cleaning

Video data quality is oftentimes influenced by changes in ambient lighting conditions. In the initial stages of dataset collection, we experimented with different lighting conditions and video editing for best results. Based on the results of these experiments, we automated the process of changing brightness and sharpness of each frame to the most optimum values for pose estimation. This gave us better results in our subsequent steps.

The data underwent a cleaning process where we manually removed videos where the player had not executed any stroke, or said player executed an ambiguous/different stroke than the labelled stroke.

### 3.4. Data diversity

Generalisation is an important aspect for developing scalable models. To recognize strokes executed by various table tennis players, a diversity in the players of our training set was necessary. Players were picked in a manner such that there was a variety in age, height, dominant hand, and years of experience playing table tennis. Diversity observed in these features ensures a range of variation in the forward stroke and follow through stages of a stroke, thus improving the generalisation of the network.

Fig. 4 shows the variation in the age of table tennis players included in the training dataset. The ages have a distribution from 10 years all the way up to 32 years of age. This provides an inherent variation in height as well. As the distribution indicates, there are multiple players in the 16 to

22 years of age category. This is because a large variation in height can be found in this age distribution owing to different timings in growth spurts.

A conscious effort was made to include a diversity in the years of experience each player has in our dataset. This factor is almost always in direct correlation with the perfection in execution of different strokes. This means that the timing of each stroke, the speed of movement, and the ROM (range of motion) will differ among players. Fig. 4 also shows the distribution of years of experience of the players in the dataset, which varies from 4 to 17 years.

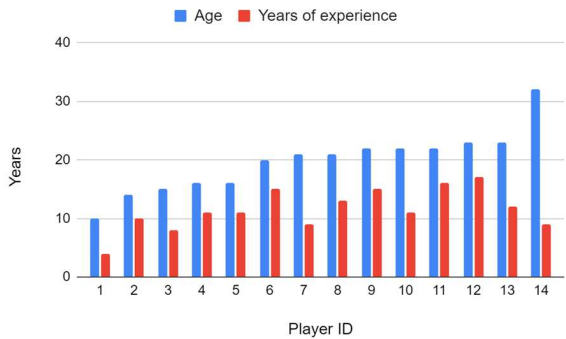

Figure 4: Distribution of age and table tennis experience in years of the players in our dataset.

## 4. Proposed method

### 4.1. Pose estimation

The collected video data has many varying factors such as player height, player gender, racket color, table color, and the general environment in which the video was recorded. Extracting features that are as independent from these factors as possible provides the potential to make accurate predictions in previously unseen or largely varying environments. Hence, we make use of two-dimensional (2D) human-body pose estimation to extract the most relevant features.

The only relevant moving parts in a given video sequence are the parts of the body of the player in reference. Using 2D pose estimation on each frame of a video sequence, we obtain a skeleton of the player's position at each time step. This effectively provides us with a trend in motion of the table tennis player across all the frames of a given video. As an initial step, a human detection algorithm is run to minimize computation costs for pose estimation. To perform this, we have used Single Shot Multibox Detector (SSD) [8] to help identify the position of the table tennis player as quickly as possible with a fairly high accuracy. Once the player's position has been localised, only this particular part of the frame is then passed through the High-Resolution Net (HRNet) [9] model to finally give us the player's pose in the corresponding frame.

### 4.2. Data preparation

Features extracted from each frame via pose estimation tend to be noisy in nature. Table tennis strokes executed by players are very quick and hence, result in slightly blurred frames in a video. This results in pose estimation mistaking different parts of the body, jittery joint localization, as well as swapping of right and left joints for certain parts of the body. To minimize this noise in the data, a Savitzky-Golay filter [10] was used. This is discussed further in section 5.3. In order to find the most suitable sample length to include as well as the most suitable order of the polynomial used to fit the samples, we mapped the extracted features onto a corresponding player over multiple test videos to look for acceptable noise reduction. It was found that the most suitable window length for the filter was 13 and the most suitable order of the polynomial was 2 (Eq. 1).

$$y_t = \frac{1}{143}[-11x_{t-6} + 9x_{t-4} + 16x_{t-3} + 21x_{t-2} \\ + 24x_{t-1} + 25x_t + 24x_{t+1} + 21x_{t+2} \\ + 16x_{t+3} + 9x_{t+4} \\ - 11x_{t+6}] \qquad (1)$$

Once smoothening of the data was accomplished, data collected from left handers was prepared to be input into our model. To simplify our further stroke recognition models, we used only the temporal pose data of right handed players and not left handed. To be able to input left handed players' data, the coordinates were flipped along the x-axis by subtracting the original x-coordinate from the width of the video resolution. To verify our results, we flipped each frame of a left handed players' video along the horizontal axis and mapped the new coordinates onto each frame.

Normalizing the input data i.e., the X and Y coordinates of the pose estimation output with respect to the video resolution, provided us with poor results. Hence, no normalization was done prior to routing the pose estimation data obtained from HRNet into our classification model.

### 4.3. Stroke recognition

In table tennis, every player has a different technique to execute certain strokes. Despite this variation, the four points of the body that show similar trends in motion between players executing the same stroke are the wrist, elbow, and the two shoulders. These four points of the body are the most necessary features in identifying a stroke being played. Thus, from the pose estimation data obtained from each video, we took the smoothened coordinates of these 4 joints (x, y coordinates for each joint) resulting in 8 features per time step.

Since input time steps vary highly with different speeds in stroke execution, we chose 100 time steps as our standard input time step size. Each of these time steps had 8 features that consisted of the coordinates of the 4 joints. In the event that the input time series data had fewer than 100 time steps, the input sequence was padded with zeros. This input sequence consisting of 100 time steps corresponding to a stroke X was fed into stroke recognition models which then classified it as X.

The methods discussed in sections 4.4 and 4.5 use a training-validation split of 95%-5% of the dataset collected.

### 4.4. Machine learning approach

Our dataset consists of multiple videos, all of 100 time steps and 8 features per time step. To input this data into a machine learning model, we first flattened this data making it 800 (100x8) features per video in the dataset. The flattened representation eliminates the temporal dimension of the pose estimation data. This data was then fit using different machine learning models adopting different algorithms. Tweaking the hyperparameters for each model, we present the best results obtained in Tab. 3.

|   | Model | Validation Accuracy (%) |
|---|---|---|
| 1 | Random Forest (trees=21) | 96.20 |
| 2 | K-Nearest Neighbours (k=3) | 92.40 |
| 3 | Decision Tree (max depth=15) | 90.32 |
| 4 | XGBoost - Random Forest | 91.68 |
| 5 | **Support Vector Machine (SVM) (RBF kernel, c=800)** | **98.37** |
| 6 | XGBoost | 98.10 |

Table 3: Comparison of machine learning models.

We tested and compared multiple machine learning models, tweaking their hyperparameters to obtain optimal training and validation results. Using the Random Forest model with more number of trees led to overfitting. The best results were obtained when the number of trees in the model were limited to 21. The KNN (K-Nearest Neighbours) model with k=3 provided the best results. Increasing this k value resulted in decreasing accuracy, hence a low k value was chosen. With a decision tree model, when the depth of the tree was not limited, it overfit onto the training data. Decreasing the depth yielded lower accuracies. A max depth of 15 for the decision tree model provided the best training and validation results, although the variance was quite high. Using an XGBoost model with Random Forest produced poorer results than an XGBoost without using Random Forest. A multi-class Support Vector Machine (SVM) model using the Radial Basis Function (RBF) kernel was also used to classify the strokes.

A one-vs-rest approach was adopted, and the normalization constant with the value c=800 yielded the best validation accuracy.

Hence, we conclude that the SVM model yields the best results since we obtained accuracies with lower variance and an acceptable bias.

### 4.5. Deep learning approach

In our deep learning approach, we use the data obtained as time series data for each video. The 100-frame input was fed into a neural network model that then classified the input stroke into one of eleven classes. Here, we compare 3 main architectures for time series analysis, an LSTM model [11], a TCN model [12], and a combined TCN + LSTM model and the overview as presented in Tab. 4. Regardless of the main architecture, the last two layers always consisted of a fully connected layer with a ReLU activation, followed by a fully connected layer with a Softmax activation for classification.

| Model | LSTM | TCN+LSTM | **TCN** |
|---|---|---|---|
| Parameters (Trainable) | 303,259 | 353,243 | **206,299** |
| Validation Accuracy (%) | 98.46 | 99.37 | **99.37** |
| Inference Time (s) [per 20000 strokes] | 2.12 | 2.61 | **0.79** |

Table 4: Comparison of deep learning models. The inference time is evaluated using an NVIDIA RTX 3070 GPU with 8GB VRAM.

Using only LSTM layers, a high bias and high variance in the training and validation accuracies were obtained. The best LSTM based model yielded validation accuracy of 96.43% when trained for 94 epochs. It was also observed from the loss curves of these LSTM models that they were sensitive to outliers in data. We experimented with models that adopted both, LSTM and TCN network architectures. A TCN was used to convolve the initial input data to fewer time steps comprising a higher number of features before propagating it through an LSTM network. The best TCN + LSTM model produced a maximum validation accuracy of 99.37% when trained for 93 epochs. However, the two above implementations using LSTM layers involved a large number of trainable parameters, higher training time per epoch, and greater inference time.

Temporal convolutions have been shown to outperform RNNs and LSTMs [13]. Hence, we experimented with models consisting of only temporal convolutional layers. As shown in Fig. 5, the 100-frame input is fed into a single temporal convolutional layer followed by a layer normalization. As mentioned in [14], layer normalization is most suitable for recurrent neural networks. However, here we found that layer normalization helps improve training even with temporal convolutional networks. Using batch normalization preceded by convolutions yielded sub-par results in terms of validation accuracy. This layer normalization is then followed by multiple ReLU-activated temporal convolutional layers and an intermediate layer normalization. These features were then flattened and propagated to a fully connected dense layer. Finally, the last layer of our model is responsible for classifying the input stroke into one of the 11 classes. This TCN based model provided the best accuracy of 99.37% on the validation set when trained for 99 epochs. The model used approximately two thirds of the trainable parameters and predicts in less than half the inference time as compared to the previously discussed models.

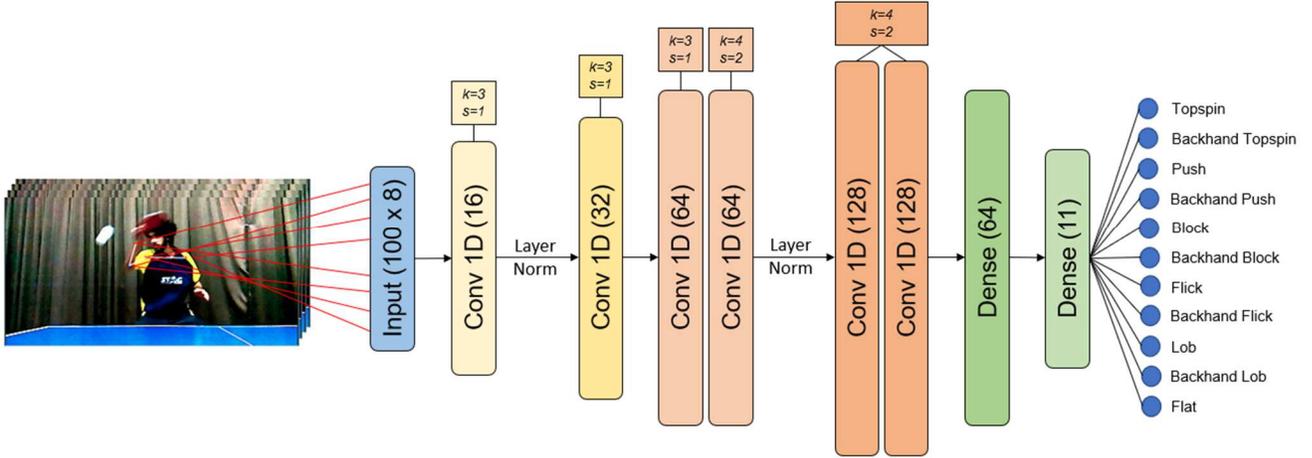

Figure 5: TCN model architecture. k – kernel size, s – stride.

## 5. Experiments on TCN model

The temporality of the pose estimation data being in its own independent dimension is not necessary for the purpose of stroke classification. This is supported by the fact that machine learning based models achieved high accuracies on flattened data as explained in section 4.4. However, deep learning approaches discussed in section 4.5 use the pose estimation data with a dedicated temporal dimension and show an improvement in validation accuracy as well as the generalisation of the model. Additionally, the temporal dimension is very essential for stroke analysis which is discussed under future scope of our work in section 7. Hence, further experiments were conducted using the deep learning based TCN model.

### 5.1. COCO vs MPII

Pose estimation is the first step in our task of stroke recognition. To ensure robust classification models, we experimented with pose estimation models trained on different pose estimation datasets. This was done by comparing the quality of pose estimation outputs for table tennis video data obtained from models trained on the COCO dataset [15], and the MPII dataset [16]. The HRNet [9] human pose estimation model was used for these comparisons.

It was found that the HRNet model trained on the COCO dataset, coupled with our TCN classification model, performed the best with a validation accuracy of 99.37%. On the other hand, the HRNet model trained on the MPII dataset, coupled with the same TCN classification model, performed worse with a validation accuracy of 82.36%. We observed that using the MPII dataset, the model has a training and validation accuracy of 80% at the 21st epoch. After this point, the training accuracy continues to steadily increase while the validation accuracy remains at 80-82%, resulting in high variance in the results.

### 5.2. Generalisation

Generalisation is of profound importance when it comes to sports analysis tasks. The measure of a model's scalability is through evaluation of its performance on data it has not been trained upon. In our case, we tested the model on the strokes executed by a player who has not been a part of the training dataset. In this way, we evaluate the behaviour of the TCN model during real world application of classifying strokes executed by a variety of players.

The strokes executed by a male right handed player with 13 years of table tennis experience (Player A) was considered first for the generalisation test. The TCN model provided an overall accuracy of 98.72% considering all 11 strokes. The confusion matrix for the predictions is depicted in Fig. 6. The player was observed to have well defined stroke actions which matched the general trend in motion for corresponding strokes from the dataset of professional players that the TCN model was trained on. This is inferred from the high generalisation accuracy for each stroke achieved by the model on his dataset.

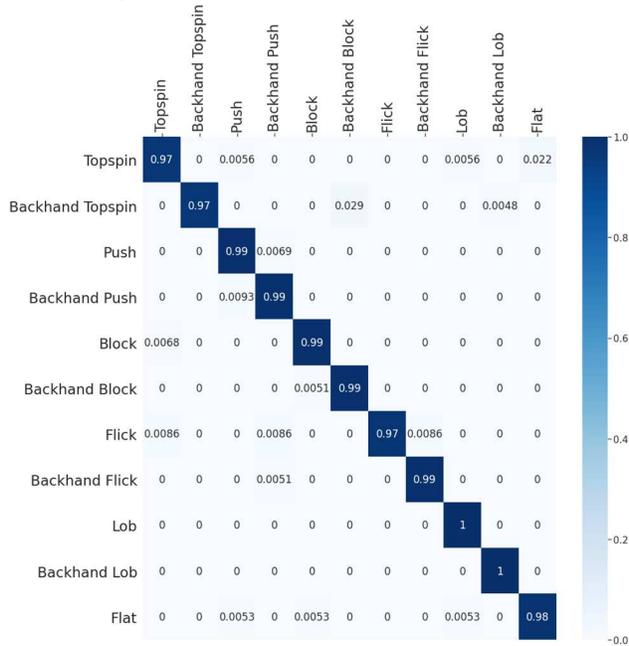

Figure 6: Confusion matrix of generalisation results for player A.

Similar generalisation test was conducted on the strokes data of a female left handed player with 10 years of experience (Player B) in table tennis. An overall accuracy of 88.81% accuracy was obtained from the TCN model. The confusion matrix of the predictions in Fig. 7(a) shows that the accuracy for her forehand push stroke was only 38%. It was observed that the player had an unorthodox forehand push action which did not agree with the general trend in motion of the same stroke from our dataset. This is a problem which could be commonly encountered in the sport of table tennis. Therefore, to make our model adapt, we accommodated these previously unseen strokes into the learning corpus. We trained the TCN model further on the data of all the 11 strokes of the player for 2 epochs, and found that the new overall accuracy was 99.22% for the same player. The confusion matrix for the model predictions is depicted in Fig. 7(b).

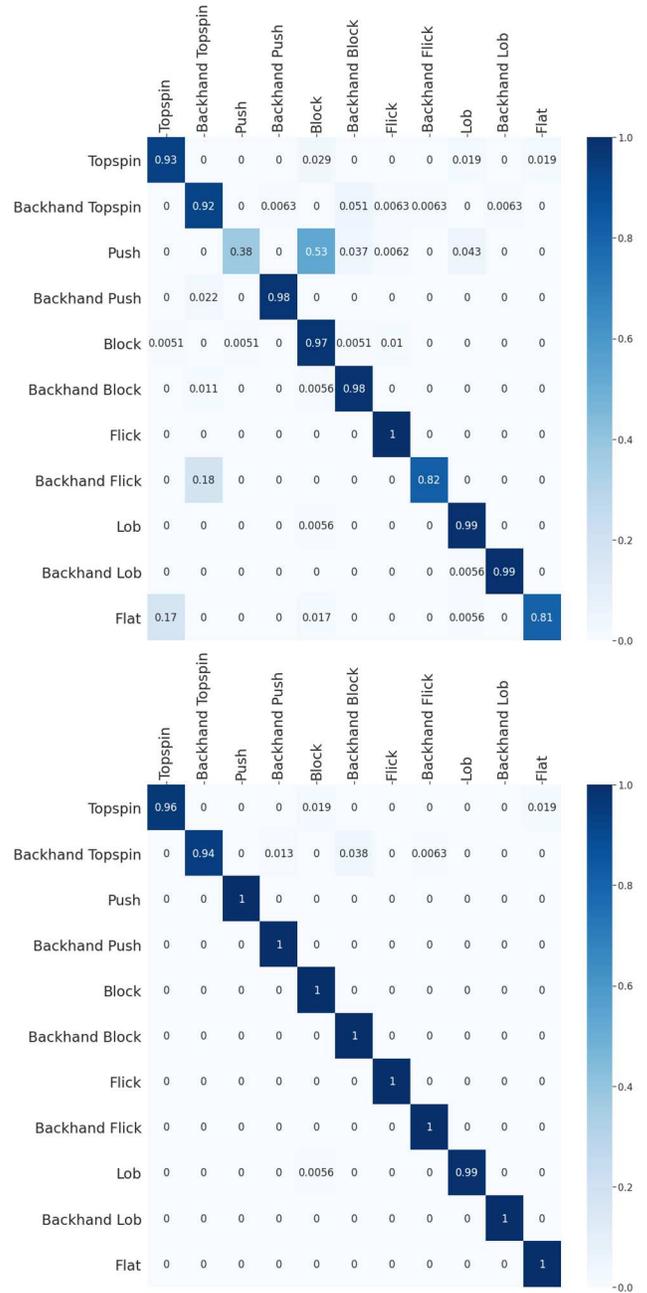

Figure 7(a) and 7(b): Confusion matrix of results for player B before training (a) and after training for 2 epochs (b).

### 5.3. Effects of Savgol filtering

Savgol filtering proved effective in reducing jittery pose estimation. This was evident upon visual inspection of pose estimation data mapped onto the video clips. One such example is shown in Fig. 8. The model shown in Fig. 5 was trained from scratch on the unfiltered pose estimation data obtained from the dataset with the same training-validation split of 95%-5%. A validation accuracy of 97.74% was obtained. Hence, Savgol filtering on the pose estimation data contributes to the improvement of validation accuracy of the model.

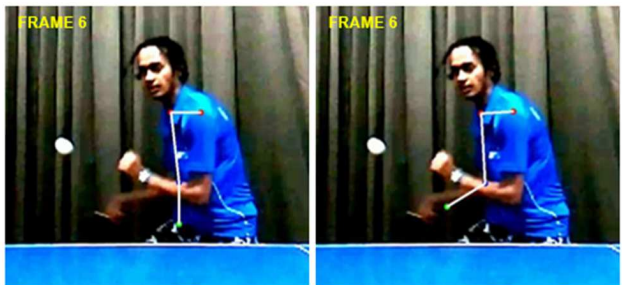

Figure 8: Pose estimation output on an intermediate frame of a forehand topspin without Savgol filtering (left) and with Savgol filtering(right).

## 6. Results and discussions

Our proposed TCN model achieves 99.37% average accuracy over the validation data with a dataset split of 95% for training and 5% for validation. As shown in Tab. 5, the model is capable of classifying each stroke executed with a high degree of accuracy leaving less room for false positives.

|    | Class            | Validation Accuracy (%) |
|----|------------------|-------------------------|
| 1  | Forehand Topspin | 99.01                   |
| 2  | Backhand Topspin | 100.00                  |
| 3  | Forehand Push    | 98.07                   |
| 4  | Backhand Push    | 100.00                  |
| 5  | Forehand Block   | 99.01                   |
| 6  | Backhand Block   | 97.03                   |
| 7  | Forehand Flick   | 100.00                  |
| 8  | Backhand Flick   | 100.00                  |
| 9  | Forehand Lob     | 100.00                  |
| 10 | Backhand Lob     | 100.00                  |
| 11 | Forehand Flat    | 100.00                  |

Table 5: Stroke-wise validation accuracy of the TCN model.

As mentioned in 5.2, a lower accuracy in classification of a particular stroke for a particular player can be attributed to an unorthodox/unconventional way of playing that particular stroke, since our model learnt the general trend of each stroke's execution across 14 players. This gives rise to the discussion of stroke error detection where a player's stroke can be compared with another player's stroke, providing differences/mismatches between the two players' strokes. Pertaining to a single player executing a single stroke continuously during multi-ball training, dips in confidence of the stroke classification could imply that the executed strokes in this time interval deviated from the player's general trend. Insights obtained from these inferences can further be used for performance monitoring during training, which may not have been obtained by conventional training methods.

## 7. Conclusion and future work

A novel method for collecting video data pertaining to table tennis has been developed for the purpose of stroke recognition. A setup with minimal interference with the gameplay and automated labelling of the feature-rich video snippets are the key features in our data collection process. The characteristic data for stroke recognition in terms of wrist, elbow and shoulder joints was extracted from the data using two-dimensional human pose estimation. Various machine learning and deep learning based models have been compared and benchmarked based on their performance on the pose estimation data. Further, the generalisation aspect of the TCN model has also been discussed.

The temporal aspect of each stroke is of utmost importance in order to analyse the quality of a stroke. Our work can be further improved upon by automating advanced methods of analysis. As discussed by Zhou Jun [17], there are 3 stages to each executed stroke. Event detection can be performed and the quality of each stage in terms of quickness and finesse can be evaluated, and inferences can be drawn. Another aspect of stroke analysis is to correct an individual player's stroke by comparing it to another world class/professional player using a neural network, providing a detailed level of correction pertaining to each stage of the stroke.

The sparingly intrusive setup for data collection as described in section 3.1, with a few modifications such as mounting the vibration sensors under the table, can be used for match analysis in competitions. Statistics for the performance of a player during a game can be provided in terms of the type and number of strokes executed, strokes which helped the player earn points and also the erroneous strokes due to which the player lost points. A player profile can be generated from the data, listing situation based as well as overall strengths and weaknesses of each player.

Lastly, lack of analytics in table tennis tends to inhibit the popularity and public outreach of such a highly competitive and interesting sport. An automated system that not only provides surface level analytics to an audience in a user-friendly manner, but also provides in-depth analytics to active professional players and coaches, can help boost the popularity of table tennis.

## 8. Acknowledgements

We thank Varun Ranganath, Shreyas Kulkarni, the SKIES TT Academy and coach Anshuman Roy, and the ACE TT Academy and coach Srivatsa Rao for their cooperation and supportive efforts making the data collection a success.